\documentclass[letterpaper,10pt,conference]{ieeeconf}
\IEEEoverridecommandlockouts
\usepackage{amsmath,amssymb,booktabs,graphicx,url,array,balance,float}
\ifdefined\pdfminorversion
\fi
\usepackage[T1]{fontenc}
\usepackage{microtype}
\title{PIER: An Evidence-Gated Execution Interface\\for Robotic Manipulation}
\author{Zoe Li$^{1}$, Anze Wang$^{2}$, Zhongyu Chen$^{3,\dagger}$, Jingran Hu$^{1,\dagger}$%
\thanks{$^{1}$University of Washington. $^{2}$Independent Researcher. $^{3}$ZJU-UIUC Institute. $^{\dagger}$Zhongyu Chen and Jingran Hu contributed equally as co-third authors.}%
}

\begin{document}
\maketitle
\begin{abstract}
Generating a plausible robot action does not establish that current observations justify its execution. Motivated by exploratory observations of high-confidence visual outputs under severe occlusion, we present PIER, an execution-authorization interface that separates evidence checks, decision provenance, and stage-scoped re-observation from hardware control. The deterministic gate evaluates declared visual and tactile inputs, while its caller maintains a budget of at most one re-observation per stage. We evaluate the implementation using 1,600 threshold-grid cases and 1,200 paired synthetic traces spanning score noise, missing tactile inputs, stale observations, and falsely reassuring scores. The finite grid yields zero declared invariant violations, and a matched Boolean baseline reproduces all non-recovery decisions. Under synthetic score noise, re-observation reduces valid-state denials from 57/120 to 18/120 while increasing invalid-state proceeds from 5/120 to 7/120. Stale and falsely reassuring inputs expose limitations that threshold checks alone cannot resolve. Exploratory visual, tactile, and robot setup records provide context but do not establish physical task performance. These results characterize an inspectable authorization interface and its input-contract limitations, without claiming superiority over equivalent rule logic, calibrated tactile accuracy, or certified physical safety.

\end{abstract}

\section{Introduction}
Action generation and execution authorization are distinct problems. This distinction motivated an exploratory integration of video perception with SAM3~\cite{sam3}, a model for prompted detection, segmentation, and tracking. A retained D405/SAM3 observation report describes severe-occlusion cases with maximum confidence 0.926--0.938 in which a downstream visual authorization rule nevertheless continued. The model produces perceptual outputs; the rule, not SAM3 itself, makes the continuation decision. Complete inference configurations and raw outputs were not recovered, so this observation motivates our question rather than serving as verified evidence of method effectiveness.

The question is whether a candidate action is supported by the observations required at its execution boundary. High model confidence is not itself a task-specific sufficiency certificate. Additional modalities may help, but missing, stale, or falsely reassuring tactile evidence creates its own failure modes. Re-observation can restore availability while also giving a noisy false positive another opportunity to pass. These trade-offs require explicit testing rather than an assumption that more modalities or retries always improve safety.

PIER emits \emph{proceed}, \emph{re-observe}, or \emph{safe stop}, retaining evidence identifiers and decision reasons. A separate adapter and operator control physical execution. We position this work as an authorization-interface and software-verification study. Its contributions are: (1) an implemented scalar-evidence interface, trace session with caller-maintained per-stage retry state, and separate offline evidence-bundle validator; (2) reproducible finite checks and a matched synthetic stress suite that expose decision equivalence, noisy retry trade-offs, and input-contract gaps; and (3) an explicit separation of verified software behavior from exploratory sensing and setup observations. The threshold predicate is not a new control algorithm, and physical Touch/Insert performance remains unevaluated.

\section{Related Work}
\subsection{Runtime assurance and authorization}
Simplex separates advanced control from a backup controller and manages control authority to maintain safety. Black-Box Simplex uses runtime checks without requiring controller internals and establishes safety under its stated assumptions~\cite{blackboxsimplex}. PIER addresses a narrower interface: declared perceptual checks, evidence IDs, decision reasons, and per-stage re-observation accounting. It neither implements a verified backup controller nor proves invariant preservation for robot dynamics. Its finite checks and traces should therefore not be equated with runtime-assurance guarantees. Ordinary Boolean logic can reproduce its predicate; the evaluation characterizes this implementation and its limitations.

\subsection{Cable manipulation and visuomotor policies}
Cable spatial representations~\cite{jin2022spatial}, hierarchical imitation learning~\cite{chen2023hierarchical}, and tactile-driven primitives~\cite{wilson2023cable} motivate retaining route progress and local contact state. Diffusion policies model action distributions for visuomotor control~\cite{chi2023diffusion}. PIER does not replace these policies; it defines an inspectable decision boundary around proposed actions. The present experiments do not compare policy learning or physical manipulation performance against these methods.

\subsection{Tactile sensing and multimodal data}
Optical tactile sensing, including DIGIT~\cite{lambeta2020digit}, makes contact observations available at the end effector. WireFishing-M provides multimodal cable-insertion data~\cite{zhou2025wirefishing}, and reactive visual--tactile policies exploit complementary feedback rates~\cite{xue2025reactive}. Our focus is how evidence availability affects authorization, rather than learning tactile features. In particular, the observed Xense image-difference signal is not equated with calibrated force or contact probability. Public datasets support interface checks only to the extent actually audited.

\begin{figure*}[t]
\centering
\includegraphics[width=\textwidth]{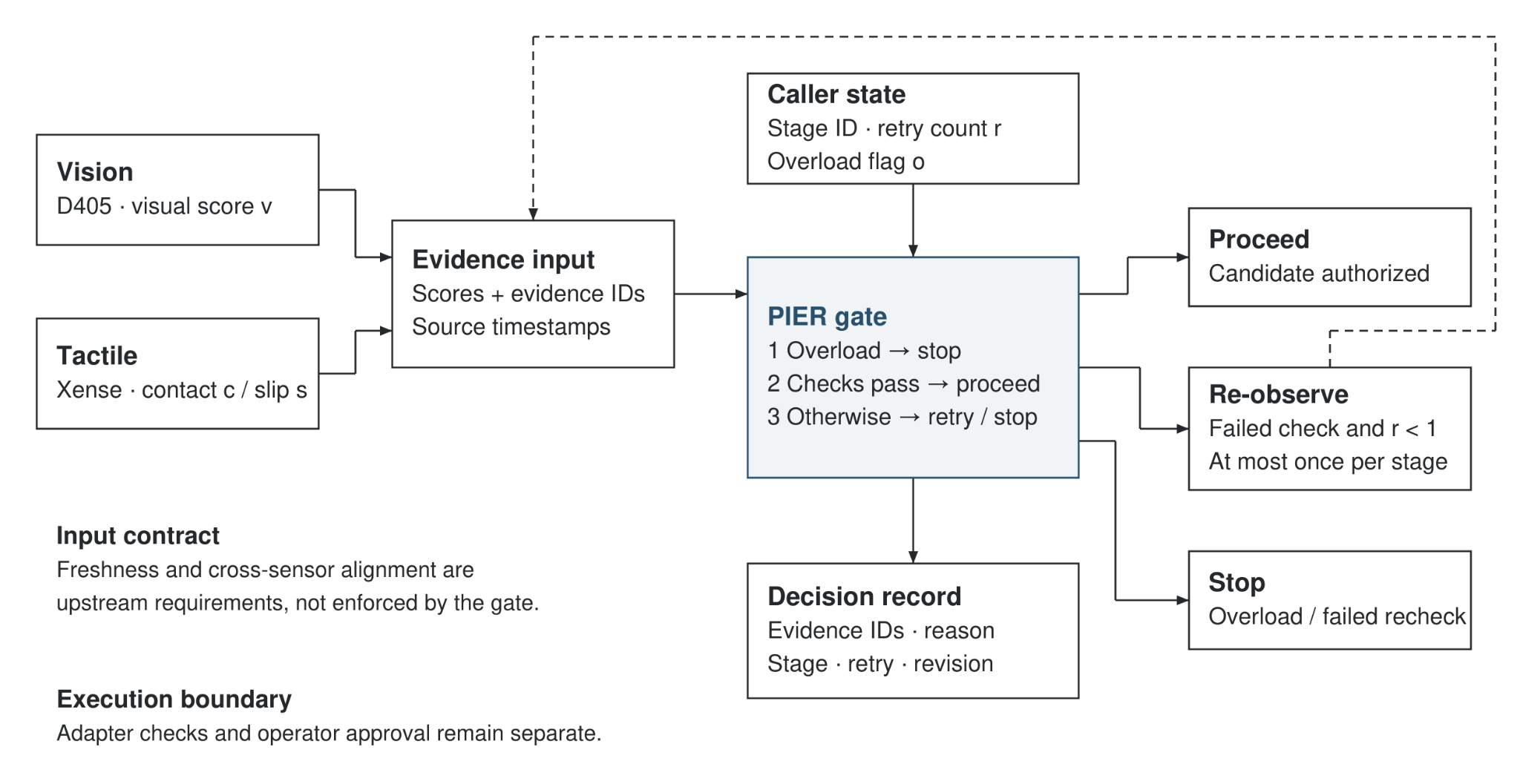}
\caption{PIER recovery-mode authorization and external execution boundary. Source timestamps do not imply enforced freshness or cross-sensor alignment. Re-observation is a software request, limited to at most once per stable stage ID within one session; the caller maintains the count. The record associates evidence, stage, and decision reason; software revision is package-level provenance. Proceed does not bypass adapter checks or operator approval.}
\label{fig:architecture}
\end{figure*}
\section{Representation and Execution Boundary}
\subsection{Episode and evidence contract}
An evidence record contains a timestamp $t$, scalar visual confidence $v$, contact confidence $c$, slip score $s$, overload flag $o$, and evidence IDs $E$. Missing tactile values remain missing; non-missing scores must be finite and in $[0,1]$. These names do not establish calibrated probabilities or physical risk measurements. The implemented call boundary is
\begin{equation}
(\mathrm{stage\_id},\ m,\ r,\ z),\qquad z=(t,v,c,s,o,E),
\end{equation}
where $m$ is the verification mode and $r$ is the caller-supplied retry count. The scalar gate uses $m,r,v,c,s,o,E$; the trace session uses the stage identifier and timestamp. Route or scene annotations can describe the experiment but do not enter the inspected predicate: there is no route-progress-dependent authorization claim.

Temporal matching and freshness are required upstream contracts. The scalar gate neither compares observations with wall-clock time nor checks cross-sensor skew. The trace session only enforces strictly increasing record timestamps. That order check does not establish that observations are recent or mutually synchronized. The 100-ms freshness wrapper evaluated below is an experimental software variant, not a live alignment implementation.

\subsection{Gate semantics}
Overload has priority and requests a stop in every mode. Otherwise, the direct baseline proceeds, the vision-only baseline requires $v_t\geq\tau_v$, and tactile-aware modes require
\begin{equation}
v_t\geq\tau_v\ \land\ c_t\neq\varnothing\ \land\ s_t\neq\varnothing
\ \land\ c_t\geq\tau_c\ \land\ s_t\leq\tau_s.
\end{equation}
The default software thresholds are $(\tau_v,\tau_c,\tau_s)=(0.70,0.65,0.35)$. A failed check stops in non-recovery modes. In recovery mode, it requests re-observation if the supplied retry count is below one; otherwise it stops. Overload never enters that branch.

\textbf{State lifetime.} A trace session initializes an empty retry map $R$ and last timestamp $t_{\rm last}=-1$. The following operations summarize the inspected implementation; recording completes before the decision is returned.
\begin{quote}\small\ttfamily
require nonempty stage\_id; $t>t_{\rm last}$\\
r := R.get(stage\_id, 0)\\
d := gate.evaluate(mode, z, retries\_used=r)\\
R[stage\_id] := r + 1 if d is RETRY, else r\\
$t_{\rm last}$ := t; append and flush trace\\
return d
\end{quote}
A previously unseen stage starts at zero; revisiting the same stage retains its count. A new session resets in-memory counts and does not reconstruct them from an existing trace file. Stage sequencing, persistent budget restoration, and an episode-wide cap are not enforced. Thus the demonstrated bound is at most one re-observation per stage per session, assuming stable stage IDs and no reset. The gate cannot itself prevent a caller from resetting or renaming stages.

\subsection{What the authorization layer contributes}
The threshold predicate is not a new decision algorithm: an ordinary finite-state executor can implement the same logic. PIER's contribution is an explicit interface joining evidence identity, modality availability, stage-scoped retry accounting, and decision provenance at the command boundary. We evaluate the predicate and its input-contract failures separately from that systems contribution. Traceability supports inspection; its effect on diagnosis time or operator performance has not been measured. We claim neither algorithmic superiority to a matched state machine nor a formal safety guarantee.

\subsection{Offline evidence-bundle validation}
A separate implemented contract validator links capture receipts, calibration, route graphs, tracks, events, evaluations, and episode records. It checks cross-file episode identifiers, referenced route nodes, ordered tracks, coordinate consistency, event-derived metrics, and declared artifact byte counts and SHA-256 digests. These are offline integrity/consistency checks, not evidence that the underlying measurements or outcome labels are true. Route geometry participates in this validator, not in the scalar authorization gate. The lightweight schema checker supports the subset used by the supplied schemas, not arbitrary JSON Schema compliance.

\subsection{Hardware and calibration boundary}
The gate itself sends no hardware commands. Its stop output withholds a candidate action; it is not a demonstrated emergency-stop latency guarantee. A reviewed adapter must separately check controller state, workspace clearance, and operator authorization. For the current modified-Xense fixed-tool setup, the operating contract keeps the gripper idle, permits zero gripper position commands, prohibits automatic physical retries, and requires separate approval for every motion-capable command. Software re-observation results do not relax this contract.

Held-out contact, slip, and load calibration remains required. The Xense record does not validate force/contact-area fusion. Sensor mappings, timing tolerances, and thresholds must be documented and frozen before physical evaluation; a failed trial must not be made to pass by relaxing thresholds.

\section{Evaluation Design and Evidence Provenance}
The primary evidence comprises an offline 1,600-case gate grid, 20 mode--scenario initial responses, a five-event trace rehearsal, and a 1,200-trace paired synthetic stress test. These runs use the supplied pure-software implementation and issue zero hardware commands. Thresholds are frozen at their existing software defaults. We distinguish invalid-state proceeds from valid-state denials and retain both outcomes rather than optimizing only one.

Exploratory records are considered separately in Section VI: a historical 60-execution software summary, a visual pilot, tactile statistics, stream coverage, and two stopped robot qualification attempts. Their missing primary artifacts prevent independent reproduction; they are not used to establish the interface's effectiveness. Counts of frames, messages, and setup attempts are never pooled as manipulation trials. No independently reviewed end-to-end Touch/Insert result is available.

The synthetic conditions are designed probes, not representative deployment samples. Reported fractions and differences are descriptive; deterministic repeated inputs do not provide independent statistical evidence, and no significance or deployment-risk claim is made.

For reproducibility, the accompanying ancillary reproducibility bundle contains \texttt{reproduce\_software.py}, \texttt{stress\_authorization.py}, and an \texttt{ARTIFACT\_MAP.md} identifying each retained output. The paper is self-contained; reproducibility materials are supplied in the arXiv ancillary files.

\section{Software Results}
\subsection{Finite invariant verification}
A fresh offline rerun evaluates the original grid: $v\in\{0,.69,.70,1\}$, $c\in\{\varnothing,0,.64,.65,1\}$, $s\in\{\varnothing,0,.35,.36,1\}$, two overload states, two retry counts, and four modes. The product is 1,600 cases. The checks test overload priority, direct-mode behavior, required modality checks, retry eligibility, and preservation of evidence identifiers.

All 1,600 cases pass with zero declared violations and zero hardware commands (Table~\ref{tab:grid}). This finite grid does not cover arbitrary inputs, caller misuse, upstream timing, robot dynamics, or asynchronous control; it is not a general safety proof.

\begin{table}[t]
\caption{Bounded software verification (fresh offline rerun).}
\label{tab:grid}\centering\small
\begin{tabular}{lr}\toprule
Check & Result\\\midrule
Enumerated threshold-grid cases & 1,600\\
Declared invariant violations & 0\\
Hardware commands & 0\\
Finite-grid verdict & PASS\\\bottomrule
\end{tabular}
\end{table}

\begin{table}[t]
\caption{Rerun initial fault responses. P/R/S: proceed/re-observe/stop. These checks validate declared logic, not sensor accuracy.}
\label{tab:fault}\centering\small
\setlength{\tabcolsep}{3pt}
\begin{tabular}{lccccc}\toprule
Mode & Nominal & Vision & Contact & Slip & Overload\\
 & & loss & missing & & \\\midrule
Direct & P&P&P&P&S\\
Vision only & P&S&P&P&S\\
V--T & P&S&S&S&S\\
V--T + re-observe & P&R&R&R&S\\\bottomrule
\end{tabular}
\end{table}

\subsection{Single-boundary fault response and trace rehearsal}
The rerun reproduces the initial decisions in Table~\ref{tab:fault}. Vision-only execution ignores missing-contact and slip variables by design, whereas tactile-aware execution uses them. This is a controlled observation-channel ablation, not evidence that a trained tactile detector is accurate.

The single-boundary campaign supplies an ideal nominal observation after a retry. Its successful second decisions are therefore scripted recovery checks, not measured recovery efficacy. The five-event trace rerun yields R, P, R, P, S with evidence IDs, stage IDs, timestamps, reasons, and retry counts retained, and zero hardware commands. It demonstrates trace serialization and stage-scoped accounting; it is not full robot replay.

\begin{table}[t]
\caption{Concrete trace case: two records from the same route stage. IDs identify synthetic evidence; no sensor files or robot motion are implied.}
\label{tab:tracecase}\centering\small
\setlength{\tabcolsep}{3pt}
\begin{tabular}{lcc}\toprule
Field & Event 3 & Event 4\\\midrule
Timestamp (ns) & 1,040,000,000 & 1,060,000,000\\
Stage & route & route\\
Evidence IDs & rgb-003, tac-003 & rgb-004, tac-004\\
$(v,c,s)$ & $(.90,.91,.82)$ & $(.91,.92,.07)$\\
Retries before / after & 0 / 1 & 1 / 1\\
Decision & RETRY & PROCEED\\
Reason & slip\_detected & visual\_tactile\_confirmed\\\bottomrule
\end{tabular}
\end{table}
Table~\ref{tab:tracecase} links a rejected observation to its replacement through a shared stage and explicit budget state. The recorded slip score explains the first decision; the replacement passes while preserving the consumed retry. Event 1 separately consumes the approach-stage budget, so the five-event session contains two retries in total. This is a worked inspection example, not evidence of faster diagnosis or improved operator performance.

\subsection{Contract fixture verification}
Nine existing contract tests pass in an offline rerun. They cover the valid bundle and independent episode checks, missing-episode rejection, required policy and coordinate-schema checks, schema declarations, dataset registry fields, and text/binary fingerprint behavior. The synthetic fixture contains nine track records and six events, with six artifact files verified across nine schema documents. This adds concrete evidence of an implemented bundle contract; it is not a physical dataset or an exhaustive corruption-resilience evaluation. The exact fixture, checker, tests, and output are included under \texttt{contract\_validation/} in the ancillary reproducibility bundle.

\subsection{Paired input-contract stress test}
To move beyond fault cases that merely mirror the checks, we run a fully retained synthetic test with noise, missing readings, stale evidence, and falsely reassuring scores (Table~\ref{tab:stress}). Five conditions each contain 120 valid and 120 invalid latent states; invalid states split equally between absent contact and slip. All policies receive the same two pre-generated readings per trace, with the latent state held fixed. This gives 1,200 paired traces and 6,000 policy evaluations, not physical episodes or independent policy samples. The generator, configuration, observations, and decisions are retained; seed 20260917 is fixed before this run and thresholds are unchanged.

Nominal scores are $(v,c,s)=(.85,.85,.10)$; absent contact sets $c=.30$, and slip sets $s=.70$. The noise condition adds independent Gaussian noise with standard deviation .20 to each scalar and clips to $[0,1]$. Dropout removes both tactile values with probability .35 per reading. Stale and false-confidence conditions supply reassuring $(.95,.90,.05)$ scores for both valid and invalid states; their ages are respectively 500 and 10 ms. These deliberately specified errors are not fitted SAM3/Xense sensor models. Clean, stale, and false-confidence counts include repeated deterministic inputs; they are diagnostic coverage, not additional independent evidence.

\begin{table*}[t]
\caption{Paired synthetic stress test. Each cell is invalid-state proceeds / valid-state denials, with separate denominators of 120 each. R: one re-observation; F: experimental freshness wrapper. All policies share inputs. No rates are interpreted as deployment risk.}
\label{tab:stress}\centering\small
\begin{tabular}{lrrrr}\toprule
Condition & Vision only & V--T & V--T + R & F + V--T + R\\\midrule
Clean & 120 / 0 & 0 / 0 & 0 / 0 & 0 / 0\\
Gaussian score noise & 97 / 31 & 5 / 57 & 7 / 18 & 7 / 18\\
Tactile dropout & 120 / 0 & 0 / 42 & 0 / 18 & 0 / 18\\
Stale reassuring input & 120 / 0 & 120 / 0 & 120 / 0 & 0 / 120\\
Fresh falsely reassuring input & 120 / 0 & 120 / 0 & 120 / 0 & 120 / 0\\\bottomrule
\end{tabular}
\end{table*}

A separately written Boolean baseline matches V--T on all 1,200 traces, confirming decision equivalence rather than superiority over ordinary rule logic. The recovery ablation uses a second noisy/missing reading, never an ideal replacement. Under noise, it reduces valid-state denials from 57/120 to 18/120 but increases invalid-state proceeds from 5/120 to 7/120: a second chance can admit a false positive. Under dropout it reduces denials from 42/120 to 18/120 without additional invalid proceeds in this run.

An experimental wrapper rejects input older than 100 ms before gate evaluation. It rejects all 500-ms records, blocking 120 invalid states while also denying all 120 valid states. This bound is a synthetic test setting, not a calibrated or deployed hardware threshold. The wrapper does not validate cross-sensor synchronization. All methods admit fresh falsely reassuring inputs; freshness cannot repair semantic miscalibration. These counterexamples bound the system claim and motivate independent evidence calibration rather than tuning thresholds to these cases.

\begin{table*}[t]
\caption{Consolidated evidence ledger. Different units and evidence classes are never pooled as robot trials. ``Reported'' indicates retained manuscript/transcript results, not a fresh raw-data rerun.}
\label{tab:ledger}\centering\small
\begin{tabular}{p{.19\textwidth}p{.28\textwidth}p{.43\textwidth}}\toprule
Evidence & Scope and provenance & Supported interpretation and limit\\\midrule
Finite gate grid & 1,600 cases; freshly rerun & Declared implementation invariants; not universal or physical safety.\\
Synthetic stress & 1,200 paired traces; 6,000 policy evaluations; new & Noise/retry trade-off, stale-input and false-confidence limits; no empirical sensor model.\\
Contract fixture & Nine tests; one synthetic bundle; rerun & Cross-file and artifact integrity checks; no physical truth or general schema-compliance guarantee.\\
Fault/trace checks & 20 mode--scenario cases; five trace events; rerun & Initial decision semantics and stage-level retry accounting; no robot replay.\\
Closed-loop software & 60 executions / 20 matched seeds; reported & Historical context only; simulator and episode outputs unavailable.\\
D405/SAM3 pilot & Nine observations; exploratory summary & Motivating observation only; rule configuration and raw outputs unavailable.\\
Xense observation & 1,201 frames; reported statistics and plot & Signal observability and baseline behavior; no contact/force accuracy.\\
Teleoperation bag & 198.99 s / 300,049 messages; reported & Heterogeneous stream coverage; no task-success or full-replay claim.\\
Public-data audits & 34,350 rows; 100-row preview; two schema/metadata profiles & Format and clock compatibility; no PIER route labels or reliability result.\\
WidowX qualification & Two reported attempts; raw JSON unavailable & Both stopped; no passed 5-mm qualification or Touch/Insert result.\\\bottomrule
\end{tabular}
\end{table*}
\section{Exploratory Records and Data Portability}
The following records provide motivation and implementation context. Their reported measurements were not regenerated from primary sensor or robot data. They do not add physical task successes to the software evaluation.

\subsection{Historical software summary}
A retained summary describes 20 matched scenario seeds under three modes (60 executions): incorrect continuation was reported as 8/20 for vision-only and 0/20 for each tactile-aware mode, with four transient-loss recoveries after re-observation. The original simulator and episode outputs are unavailable. These counts are historical context only, excluded from the primary effectiveness claims and not pooled with the reproducible stress test. Their uncertainty and selection process cannot be resolved from aggregate counts.

\subsection{Command-free visual stress pilot}
\begin{figure*}[t]
\centering
\includegraphics[width=\textwidth]{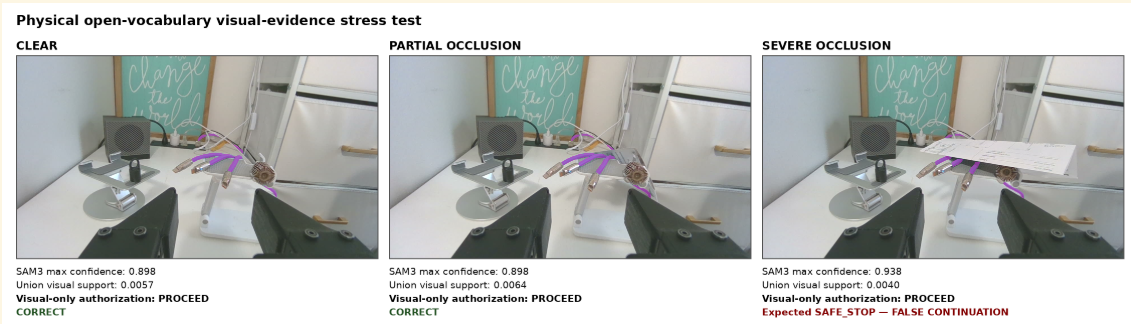}
\caption{Exploratory visual pilot: retained clear, partial-occlusion, and severe-occlusion examples. The image annotations report a downstream visual rule proceeding in all three, including severe occlusion with confidence 0.938 and an expected stop label. Numerical annotations are transcribed in Table~\ref{tab:sam3} for readability. Full configuration and primary outputs are unavailable; this is not a reproduced inference benchmark or evidence of tactile correction.}
\label{fig:sam3}
\end{figure*}
A retained pilot summary reports nine D405 observations described as held out and evaluated by a frozen visual-support rule using SAM3~\cite{sam3}. The rule agrees with the predeclared evidence-sufficiency labels on all six clear/partial observations, but incorrectly continues on all three severe-occlusion observations. Maximum SAM3 confidence for the severe cases lies between 0.926 and 0.938. These are observation-level decisions without robot commands, not nine manipulation trials.

Figure~\ref{fig:sam3} and Table~\ref{tab:sam3} make the failure visible: the severe example has higher confidence (0.938 versus 0.898) but lower reported union support (0.0040 versus 0.0057/0.0064), and still proceeds. Model confidence and visible support are not interchangeable with execution sufficiency.

\begin{table}[t]
\caption{Representative SAM3 comparison transcribed from Fig.~\ref{fig:sam3}. Confidence and support values belong to the displayed examples, not condition averages.}
\label{tab:sam3}\centering\small
\setlength{\tabcolsep}{3pt}
\begin{tabular}{lccc}\toprule
Condition & Clear & Partial & Severe\\\midrule
Maximum confidence & 0.898 & 0.898 & 0.938\\
Union visual support & 0.0057 & 0.0064 & 0.0040\\
Rule decision & Proceed & Proceed & Proceed\\
Expected decision & Proceed & Proceed & Stop\\
Agreement & Correct & Correct & Incorrect\\\bottomrule
\end{tabular}
\end{table}

Across the nine observations, agreement is 6/9 and false continuation is 3/9 overall, or 3/3 within severe occlusion. The three illustrated samples are a subset, not additional trials. This small exploratory pilot motivates a distinction between confidence and sufficiency; incomplete artifacts prevent a causal analysis of the rule failure or a reproducible model comparison. The rule was reportedly frozen, but its prompt/target specification, checkpoint, split manifest, label protocol, raw masks, and complete outputs remain unavailable. It is unclear whether maximum confidence always belongs to a task-relevant target. The displayed expected decisions are reported annotations, not independently relabeled ground truth. We retain ``union visual support'' without inventing its normalization or threshold; inference was not rerun.

\subsection{Xense sensing and baseline behavior}
The physical Xense record comprises 1,201 frames at approximately 30 Hz across operator-prompted no-contact, mixed gentle-touch/release, and release phases. Table~\ref{tab:tactile} lists the reported phase statistics. The mean absolute pixel-difference (MAD) proxy is a descriptive image signal, not calibrated normal force or contact confidence.
\begin{table}[t]
\caption{Reported Xense phase statistics. Prompt phases are not contact ground truth; frames are temporally correlated.}
\label{tab:tactile}\centering\small
\begin{tabular}{lrr}\toprule
Prompt phase & Frames & Median MAD\\\midrule
No contact (full phase) & 300 & 2.6522\\
Mixed touch/release & 660 & 2.7168\\
Release & 241 & 2.7229\\\bottomrule
\end{tabular}
\end{table}
After excluding the first second of reference-frame initialization, the no-contact median is 2.6562. The mixed-phase and release medians exceed that value by 0.0606 and 0.0667, respectively. Release remains elevated rather than returning immediately to baseline. This pattern is consistent with drift or hysteresis, but the record alone does not isolate their physical causes. Prompt transitions do not provide exact contact onset or independently labeled release events. We therefore report neither contact accuracy nor slip-detection accuracy, and do not treat 1,201 correlated frames as 1,201 independent trials.

\begin{figure*}[t]
\centering
\includegraphics[width=.95\textwidth]{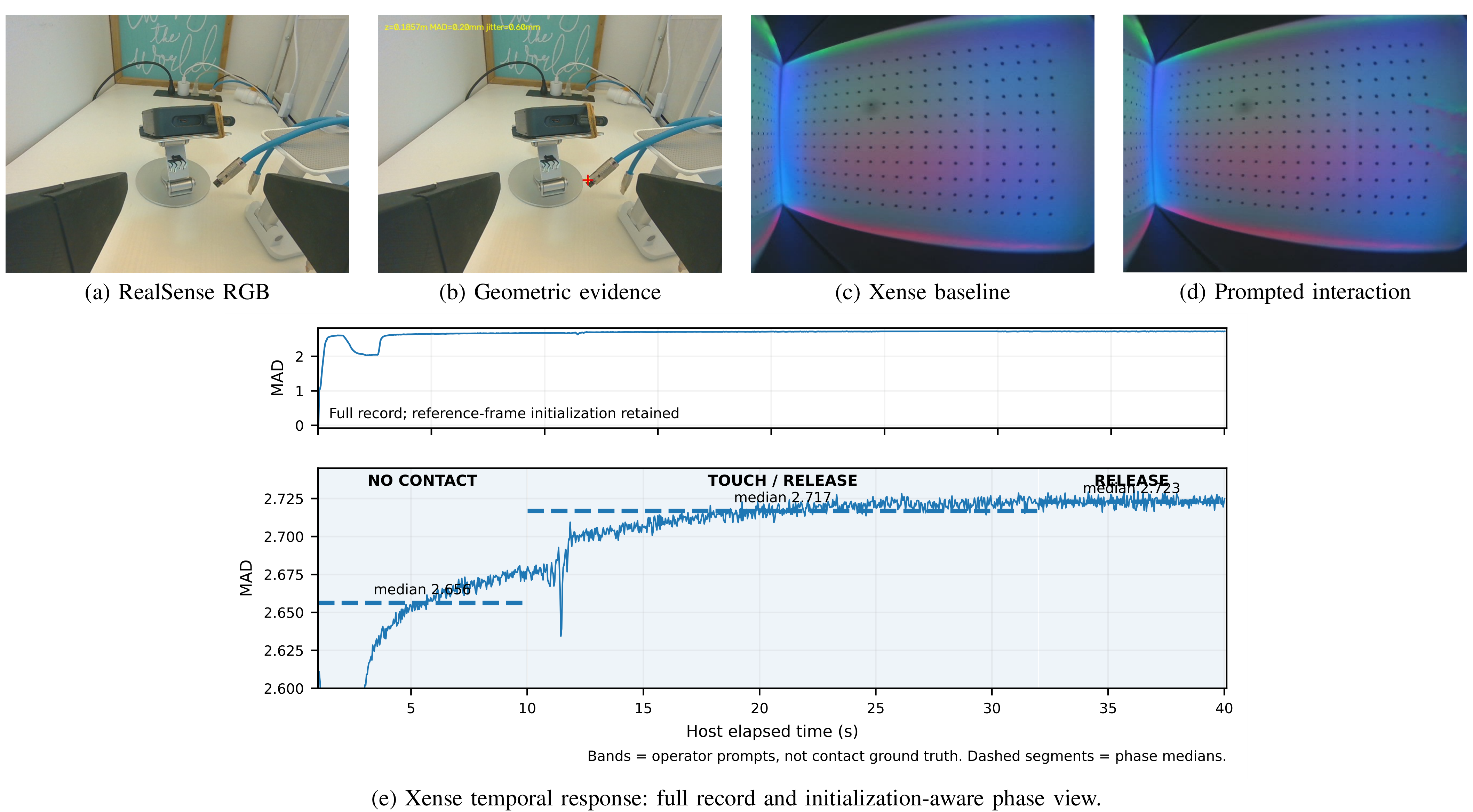}
\caption{Exploratory physical observation panel: D405 RGB/geometric views, Xense views, and the recorded MAD timeline. The original full-record and initialization-aware views are preserved without reconstructing raw samples. Shaded bands denote operator prompts, not independently verified contact states.}
\label{fig:physical}
\end{figure*}

\subsection{Real-robot stream coverage}
The retained teleoperation bag summary reports 198.99 s and 300,049 ROS2 messages. Representative counts are 2,976 RGB images, 2,988 depth images, 3,739 current-pose messages, 34,648 joint-state messages, 19,782 TCP-feedback messages, and 3,015 end-effector targets. These selected streams sum to 67,148 messages; the remaining 232,901 belong to other streams in the reported total. The subset is not presented as a complete topic inventory.

This historical bag supports stream availability, not synchronized decisions, replay, or task success. Its target messages do not establish gripper actuation in the separate current idle-gripper workflow.

\begin{figure}[t]
\centering
\includegraphics[width=.94\columnwidth]{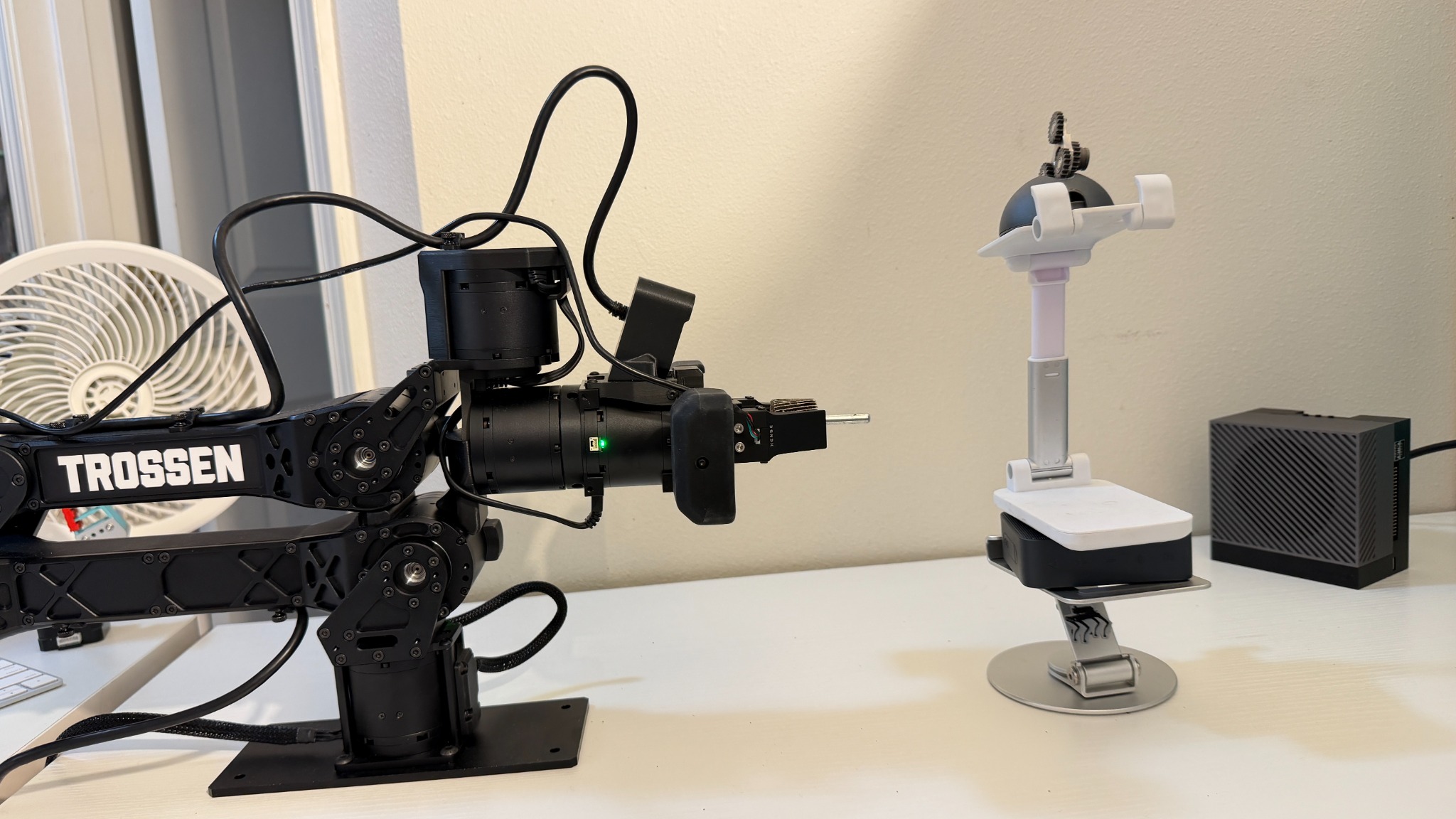}
\caption{WidowX with the modified Xense fixed tool. This setup photograph documents hardware context; it does not establish tool-to-hole alignment, calibrated geometry, or a successful insertion.}
\label{fig:setup}
\end{figure}
\subsection{Supervised fixed-tool qualification}
Figure~\ref{fig:setup} shows the retained WidowX/Xense setup. A retained experiment summary reports two fixed-tool free-space attempts (Table~\ref{tab:qualification}). Attempt 001 stops at preflight before Cartesian dispatch. Following a correction to the interpretation of the live position-tolerance field, attempt 002 dispatches one nonblocking 5-mm command. Its maximum signed progress is only 0.190 mm before the reported J2 external-effort change reaches 1.384 Nm, exceeding the frozen 1.350-Nm envelope. Neither attempt passes the requested qualification.

\begin{table}[t]
\caption{Exploratory fixed-tool qualification. Values are retained summary claims; the original JSON records were not recovered for independent verification.}
\label{tab:qualification}\centering\small
\begin{tabular}{p{.12\columnwidth}p{.76\columnwidth}}\toprule
Attempt & Reported outcome\\\midrule
001 & Preflight SAFE\_STOP; zero Cartesian dispatches.\\
002 & One 5-mm dispatch; maximum signed progress 0.190 mm; J2 effort delta 1.384 Nm exceeds 1.350 Nm; SAFE\_STOP.\\
Both & Zero gripper position/non-idle-mode commands; zero automatic returns or retries.\\\bottomrule
\end{tabular}
\end{table}

The same report records a stable negative gripper coordinate near $-3.618$ mm without a controller error. That unresolved coordinate does not authorize gripper actuation. These are reported examples of preflight and runtime gating, not independently verified stop timing or certified safety. The primary JSON files remain unavailable in the readable evidence package.

Earlier partial records retain a supervised wrist-only PASS and fixed-tool idle telemetry; a 35-segment posture plan remains an unverified candidate. Their scope is bring-up, not hand--eye/TCP accuracy or completed contact manipulation. Per-segment telemetry and operator clearance remain required, and no insertion outcome is inferred from the setup photograph.

\subsection{Public-data adapter audits}
The retrieved row-audit artifact for DistantSky's TsFile mirror~\cite{distantsky2026} reports 34,350 checked rows across 72 episodes, complete required fields, and no per-episode timeline failures. The retained audit description specifies 14-dimensional state, velocity, processed-action, and raw-action fields. Median/max source-action versus container-clock differences are 72.56/154.19 ms; raw-to-processed action skew is 2.53/10.77 ms. These two clock comparisons must not be conflated with live sensor-alignment error.

The community AIC sample~\cite{aicsample2026} has a bounded first-100-row check with no reported dimension/timeline failure; it is a simulation-derived preview. WireFishing-M~\cite{zhou2025wirefishing} is a published-schema mapping and Yaskawa~\cite{yaskawa2026} a metadata audit. The latter two were not downloaded as physical evaluation datasets in the retrieved audit. None provides PIER-specific route ground truth or a PIER task-success benchmark. Table~\ref{tab:ledger} summarizes the claim boundaries.

\section{Limitations and Next Physical Evaluation}
Gate correctness depends on calibrated, timely inputs and correct retry accounting. Thresholds cannot repair stale evidence or high-confidence visual insufficiency. Designed scenarios, correlated frames, messages, and motion attempts have distinct statistical meanings; none establishes cross-scene reliability. Several results are retained reports whose unavailable primary artifacts are identified in Table~\ref{tab:ledger} and the ancillary reproducibility bundle.

The next physical study requires validated camera-to-robot and tool-tip geometry, independent contact and insertion outcome labels, and a reviewed executor. The intended sequence is staging, command-free verification, 5-mm free-space qualification, pre-insert, touch, segmented insertion, and final visual/geometry verification. Every physical attempt, including stopped and aborted attempts, must have an append-only record; safety thresholds remain fixed and every motion requires separate operator approval. The modified gripper remains idle with zero position commands. This is a prospective protocol, not a completed experiment.

One verified success would establish feasibility. Comparative reliability additionally requires repeated trials, condition-specific denominators, independent labels, and retained failures.

\section{Conclusion}
PIER defines an inspectable execution-authorization interface with caller-maintained stage budgets and recorded evidence IDs and reasons. Finite checks reproduce the declared predicate; a matched Boolean baseline confirms that the predicate itself is not an algorithmic advance. Paired synthetic probes expose a trade-off: re-observation reduces valid-state denials but can increase invalid-state proceeds under noise. Timestamp order and high scores do not establish fresh or semantically sufficient evidence. Exploratory sensing and setup records motivate these concerns without validating physical task capability. The supported result is a reproducible characterization of the interface and its limitations, not a demonstrated insertion system or a certified safety mechanism.

\section*{Acknowledgments and AI Disclosure} The authors thank the Mechatronics, Automation, and Control Systems Laboratory (MACS Lab) at the University of Washington and Professor Xu Chen for their support of this work.\par
OpenAI ChatGPT assisted with drafting and restructuring Sections I--VIII, table and caption wording, and generating the software-only stress-test script described in Section V. The script was executed and its outputs are retained in the ancillary reproducibility bundle. AI assistance did not generate physical measurements or replace hardware experiments. Authors remain responsible for checking all claims, references, code, and reported results.

\balance

\end{document}